\documentclass[manuscript,screen,nonacm]{acmart}

\usepackage{fancyhdr}
\AtBeginDocument{%
    \addtolength{\footskip}{2.0\baselineskip}%
    \fancyfoot[L]{\textit{\textbf{Preprint --- final version forthcoming}}}}%

\AtBeginDocument{%
  }

\setcopyright{none} % Removes the copyright notice.
\renewcommand\footnotetextcopyrightpermission[1]{}
\usepackage{tabularx}
\usepackage{cleveref}
\usepackage[noend]{algorithm2e}
\LinesNumbered
\RestyleAlgo{ruled}
\crefname{algocf}{alg.}{algs.}
\Crefname{algocf}{Algorithm}{Algorithms}
\usepackage{bbm}
\usepackage{amsmath}

 \usepackage{enumitem}

\begin{document}

\title[Do You Trust the Process?]{Do You Trust the Process?: Modeling Institutional Trust for Community Adoption of Reinforcement Learning Policies}

%Designing Interventions to Increase Community Trust for Fair Network Outcomes

\author{Naina Balepur}
\email{nainab2@illinois.edu}
\affiliation{%
  \institution{University of Illinois Urbana-Champaign}
  \country{USA}
}

\author{Xingrui Pei}
\email{xpei5@illinois.edu}
\affiliation{%
  \institution{University of Illinois Urbana-Champaign}
  \country{USA}
}

\author{Hari Sundaram}
\email{hs1@illinois.edu}
\affiliation{%
  \institution{University of Illinois Urbana-Champaign}
  \country{USA}
  }


\begin{abstract}
Many governmental bodies are increasingly adopting AI policies for decision-making. In particular, Reinforcement Learning (RL) has been used to design policies that citizens would be expected to follow if implemented. 
Much RL work assumes that citizens do indeed follow these policies, and evaluate them with this in mind.
However, we know from prior work that without institutional trust, citizens will not follow policies put in place by governments. In this work, we develop a trust-aware RL algorithm for resource allocation in communities. We consider the case of humanitarian engineering, where the organization is aiming to distribute some technology or resource to community members. We use a Deep Deterministic Policy Gradient approach to learn a resource allocation that fits the needs of the organization. Then, we simulate resource allocation according to the learned policy, and model the changes in institutional trust of community members.
We investigate how this incorporation of institutional trust in RL algorithms affects outcomes, and ask how effectively an organization might be able to learn policies if trust values are not public. 
We find that incorporating trust into RL algorithms can lead to more successful policies, in particular, when the organization's goals are less certain. In addition, we find that more conservative trust estimates lead to increased fairness and average community trust, though organization success suffers. Finally, we explore one strategy to prevent unfair outcomes to communities. We implement a quota system put in place by an external entity which decreases the organization's utility when it does not serve enough community members.
We find this intervention can improve fairness and trust among communities in some cases, while decreasing the success of the organization. This work underscores the importance of institutional trust in algorithm design and implementation, and identifies a tension between organization success and community well-being.
\end{abstract}


\keywords{Reinforcement Learning, Resource Allocation, Institutional Trust}

\maketitle

\section{Introduction}

In recent years, reliance on algorithmic governance has increased dramatically \cite{danaher2016threat, gritsenko2022algorithmic, katzenbach2019algorithmic, issar2022algorithmic}; with it, so has the prevalence of algorithmic governance research.
In this work we focus on Reinforcement Learning (RL) algorithms, which have been used widely to learn policies in response to changing environments. In particular,  
COVID-19 policies including border control, vaccination location setting, and resource allocation \cite{bastani2021efficient,kompella2020reinforcement, kwak2021deep, padmanabhan2021reinforcement, uddin2022optimal, zong2022reinforcement}, and taxation and debt collection policies \cite{zheng2022ai, abe2010optimizing, mi2023taxai} have all been proposed through RL algorithms. 
Work in these contexts often focuses on learning a policy that will help governments achieve the optimization of some goal, which may focus on social welfare---like taxation policies to decrease economic inequity or border control to reduce the spread of COVID-19. 
% this should be more about whether agents' opinions on the governmental body using these policies can change as the policies are put in place.
% and then connect the two bodies of literature with this?

However, we know that the use of AI in these contexts can have two major issues. First, algorithms can cause harm to marginalized groups \cite{chander2016racist,angwin2022machine,whittaker2018ai}. 
Second, and the focus of this work, is that algorithmic decision-making can \textit{decrease} citizens' trust in those decisions. \citet{ingrams2022ai} find that citizens perceive algorithmic decision-makers to be less competent, honest, and benevolent than their human counterparts. Here, we focus on \textit{institutional trust}, or the belief by citizens that institutions are acting according to the expectations of the public \cite{miller1990political}.

Understanding how institutional trust of citizens is fostered and dissolved is key \cite{Alexander2018Why, Kennedy2021Trust, Gulati2019Design}. This trust can affect adoption of technology, use of services, and compliance with policies, like social distancing during COVID-19 \cite{komiak2006effects, lee2011much, caplanova2021institutional}. 
%In fact, increased institutional trust can even increase citizens' willingness to pay taxes to support the welfare state \cite{habibov2018does}. 
If the use of AI policies breeds mistrust \cite{ingrams2022ai}, and this mistrust can directly affect citizens' actions, the simulated outcomes of our policies cannot be accurate if we do not take institutional trust into account. This is especially salient as we experience a global decrease in institutional trust in Western governments \cite{Lounsbury2023The}. \textit{Much work in developing RL algorithms for governmental policy ignores this tension, and assumes that citizens follow the policies}, or in some cases \cite{zheng2022ai}, accounts for selfish behavior of agents without considering trust. 
Understanding and modeling how these algorithms are received is essential as we consider how communities react to AI decision-makers, including refusal to comply with policies. 

%The engineers are limited both in manpower and in the amount of technology that can be dispersed. The engineers use their limited time to do outreach in communities to gain the trust of community members. While doing this outreach they also spend time allocating the technology. The community members exist on a social network and their trust in the engineers is affected both by their receiving of the technology and also the time that is spent by with engineers on trust increasing outreach. The engineers are tasked with dispersing technology to maximize some utility function. This utility might be dependent on some fairness metrics, but also how many resources are leftover afterward. Engineers who did not wish to do outreach may decide that trust is not as important to them.

In this work, we focus on modeling and predicting the effect of RL policies on the institutional trust of community members; we then design RL models that will take this trust into account. 
Specifically, we consider a community network of citizens and a humanitarian engineering organization (\textit{e.g.,} Engineers Without Borders \cite{ewb}, which provides engineering solutions to under-resourced communities).  
In our model, each community member has some prior trust in the organization (due reputation and individual propensity to trust) \cite{mayer1995integrative, hancock2023and}; its goal is to allocate some technology to the community while staying within budget. 
Using a Deep Deterministic Policy Gradient (DDPG) framework, the organization implements a policy that will achieve its goals of staying under budget and equitably allocating the technology (generically, we call these resources or services). After the technology has been distributed, citizens all update their trust levels \cite{granatyr2015trust} according to the services they received and their perception of the fairness of the algorithm. 
As a citizen's trust level falls, they will begin to refuse services from the organization, which may in turn disrupt the organization's ability to distribute resources. Thus, fostering the institutional trust of community members is essential.

We develop three RL models: trust-unaware (trust is ignored), trust-aware (initial trust values of community members are known), and learned-trust (initial trust values are unknown but other signals allow for Bayesian estimates). We find that for all models, as the organization's desire to save resources increases, the RL agent is more successful, but the trust of community members is lost. When the organization attempts to equally prioritize saving resources and serving the community, the additional trust information in the trust-aware algorithm changes outcomes dramatically. 
%For high initial trust values in the community, our trust-aware algorithm outperforms trust-unaware in fairness and final community trust, but organization success suffers. The opposite is true for initially low trust values (organization success is improved but fairness and trust suffer).
Surprisingly, the learned-trust model is most able to improve fairness and trust in the community, outperforming both trust-aware and unaware. However, this again comes with a clear tradeoff that organization success suffers. We posit that Bayesian estimates of trust, as conservative predictors, produce the most fair, trust-improving policies. 
The network structure of community members also has an effect on outcomes---higher degree nodes are more likely to lose trust in the organization as they are likely to observe unfair outcomes in their larger neighborhoods.
Finally, we study one solution to unfair outcomes: placing a service quota on the organization. As expected, the quota harms the organization's success rate when the organization values saving resources over providing them. When the organization's goals are more balanced, this intervention promotes fairness and trust-improving policies.
We make two contributions:
\begin{description}[style=unboxed, leftmargin=.35cm]
    \item[Trust-Aware RL Policy: ] We are the first to propose learning a policy while incorporating trust. Prior RL literature ignores the tension between AI decision-makers and the institutional trust of those affected by algorithms. Not only does this disregard the impacts to citizens, but also leads to less accurate and informative policy simulations. In this work, we model the effect of governmental decision-making RL algorithms on the trust of citizens, and then use these trust values as inputs while we learn policies. 
    We compare these informed policies to uninformed ones, and determine under what circumstances and to what degree trust should be incorporated into these models. 
    Without this consideration, we are missing a key attribute that can help us develop more predictive and fair models.
    \item[Quantifying Community Harms: ] We quantify community harms and simulate an algorithmic solution.
    While lack of trust is partly fueled by the use of AI itself, RL policies can also cause harm to specific individuals \cite{chander2016racist,angwin2022machine,whittaker2018ai}. Humanitarian engineering organizations in particular have caused harm and promoted neocolonialism \cite{birzer2019humanitarian}, even unintentionally. In this work we quantify the harm to communities when human decision-makers do not prioritize fairness. Further, in contrast to prior work, we propose a possible algorithmic solution---service quotas placed on the organization. Understanding the harms that can come from policies is largely under-explored; to responsibly develop algorithms we must predict future harm and propose solutions.
\end{description}

Our results demonstrate the importance of institutional trust to community policies. With sufficient data, these simulations can be used as predictive tools to estimate downstream effects of implementing RL policies in communities, perhaps in conjunction with proposed solutions to decrease inequitable algorithmic outcomes, like participatory design for production of fair and efficient public decision-making algorithms \cite{danaher2017algorithmic, lee2019webuildai, veale2018fairness}.

\section{Related Work}
\textbf{Reinforcement Learning for Policy: }
Reinforcement learning has been used in many areas to design real policies.
\citet{yu2021reinforcement} survey uses of RL in the healthcare sector. In particular, \citet{deliu2023reinforcement} designs a messaging system that implements an RL algorithm to deliver personalized behavioral recommendations to users to promote wellness. COVID-19 policies including border control, vaccination location setting, and resource allocation \cite{bastani2021efficient,kompella2020reinforcement, kwak2021deep, padmanabhan2021reinforcement, uddin2022optimal, zong2022reinforcement} have also been proposed through RL algorithms.
In the climate activism space, RL techniques have been used to learn policies for governments to implement in response to rising sea levels \cite{shuvo2019scenario}. RL has also been used in areas of economic governance, like taxation and debt collection \cite{zheng2022ai, abe2010optimizing, mi2023taxai}. While some work has considered ethical implications of using RL to solve high-stakes problems \cite{greene2023taking, whittlestone2021societal, chapman2022power}, and others have studied the difficulties in implementing real-world solutions \cite{dulac2021challenges}, very little work has modeled how algorithms affect community trust \cite{balepur2024intervening}. In particular, no work to date has considered the relationship between citizen trust and effectiveness of RL policy.

\textbf{Fair Resource Allocations: }
Deciding how to fairly allocate resources is a problem with a rich body of literature in the game theoretic space \cite{maritan2017resource, brown1984concept, bower2005resource, hurwicz1973design}.
One application area is disaster relief---\citet{donmez2022fairness} survey the humanitarian logistics space, covering many methods for fair resource allocation. Prior work has developed resource allocation policies that can be adjusted to decision-makers needs \cite{hu2016bi}, while other work presents several heuristic solutions for fair distribution of food rescue resources, simultaneously ensuring efficiency \cite{nair2017fair}. Very little work has used RL methods to assign fair allocations; rather, most prior work has relied on heuristic approaches. In our work we use RL as a tool to mitigate the challenges of allocating resources within the changing dynamics of the environment. 

One metric we can use to measure the fairness of resource allocations is the Gini coefficient \cite{dorfman1979formula}. Originally used to measure economic inequity, it has since been used in various domains, including education inequality \cite{thomas2003measuring}, and even geography \cite{han2016using}. In this work, we extend the Gini coefficient to determine the inequity of service allocations.

\textbf{Institutional Trust: }
Institutional trust is the belief by citizens that institutions are acting according to the expectations of the public \cite{miller1990political}. Work has been done in sociology and business management to understand which factors can alter citizens' trust in institutions;
in particular, \citet{kaasa2022determinants} find that individuals tend to trust institutions less when there is a large power distance, meaning sense of participation and civic responsibility are important factors. \citet{PytlikZillig2017A} find experimentally that increased knowledge stabilizes institutional trust attitudes of community members. The COVID-19 pandemic also brought focus to issues of institutional trust; \citet{Jiang2022In} categorize citizens using unique trust profiles, and find correlations with compliance with governmental COVID-19 protocols. In this paper we implement many of these findings, allowing institutional trust to change over time with additional knowledge.

\textbf{Human-AI Interaction: }
Human-AI interaction is a field that encompasses how humans interface with AI. In this work, we consider human reactions when faced with AI decision-makers, specifically fluctuations in trust. 
In theoretical work, \citet{manzini2024should} investigate whether trust in AI assistants is justified, which they study as a function of competence and alignment.
In applied work, researchers study the degree of trust developers have in AI coding tools, and find that trust is dependent on the tool's capability, integrity, and benevolence \cite{wang2024investigating}. 
\citet{purves2022public} find that AI opacity leads to loss of institutional trust from citizens. 
There has also been in-depth exploration of trust in the field of human-robot interaction \cite{yagoda2012you, hancock2011meta, law2021trust, lewis2018role}; most relevant to our work, \citet{guo2021modeling} model and predict human trust in robots after they interact.  
In this work, we model institutional trust being lost or gained by citizens as a direct effect of RL algorithms. Our formulation and attempt to quantify institutional trust in this way is novel.

\section{Problem Statement}\label{sec:ps}
Consider a community network $G = (V,E)$ with community members $v \in V$. 
The set of undirected edges $E$ represents the set of active relationships maintained among individuals, along which information can flow. 

Now, consider a humanitarian engineering organization, $\mathcal{H}$, which arrives in this community. The organization is not part of the network, but is able to observe it. The amount of information that can be observed by $\mathcal{H}$ is model dependent, and summarized in \Cref{tab:models}. The goal of organization $\mathcal{H}$ is to provide some technology, like solar panels or cook stoves, to the community and distribute them equitably \cite{Ramírez2011Participative}. The organization $\mathcal{H}$ has a budget of $\rho$, which represents the amount of resources it has to distribute. Say also that $\mathcal{H}$ has a private attribute $c \in [0,1]$, the organization's willingness to decrease the quality of the service in order to stay below its resource constraint $\rho$.

Define a service to be provided by organization $\mathcal{H}$ to one community member $v$, with an associated real-valued service quality $s_v$, such that $\sum_{v\ \in V} s_v \leq \rho$. For example, providing solar panels to a community will involve some allocation of finite resources which will vary across community members.  
Each service provides some utility $u_v(s_v)$ to one individual $v$, dependent on quality of service $s_v$.  
The organization chooses how to allocate services in order to maximize its own utility metric, $\mathcal{U}$. This metric may be a function of how many people are affected by the services, the quality of the services, any remaining resources from $\rho$, and the private attribute $c$ representing the willingness of $\mathcal{H}$ to reduce the quality of services to stay under budget. Note that if $c$ is high, the formulation of this utility function could encourage the organization $\mathcal{H}$ to decrease the quality of the service (\textit{i.e.,} to cope with budget cuts).

Community members will only accept services if they trust the humanitarian engineering organization $\mathcal{H}$. Each individual $v$ begins with a baseline institutional trust level of $\tau_v$, which will be updated each iteration. The change in trust is a function of the utility $u_v$ from any services $v$ received, as well as the fairness of service utility distribution to its neighbors.
Note that $v$ cannot observe the utility of all community members $u \in V$; $v$ only has knowledge of individuals they share an edge with.
In this work we specifically study trust as a social phenomenon, and investigate how poor service quality from organizations, especially from AI decision-makers, can break trust with community members. We seek to answer the following questions:

\begin{description}[style=unboxed, leftmargin=.35cm]
    \item[RQ1: ] Suppose the institutional trust value of each community member $\tau_v$ is public. Can the humanitarian engineering organization $\mathcal{H}$ learn the optimal policy to allocate technology while minimally eroding trust? How effective is this policy when compared to one that does not incorporate trust?
    \item[RQ2: ] Suppose the institutional trust value of each community member $\tau_v$ is, instead, private. Can $\mathcal{H}$ learn these trust values while learning the policy? How does this lack of knowledge affect the success of the policy?
    %\item[RQ3: ] Consider a scenario where the organization $\mathcal{H}$ is making trade-offs that maximize their utility but harm community members. Through observation of services, can community members collectively learn some properties of the organization's utility function $\mathcal{U}$? If so, can they strategically accept or reject services to alter their outcomes?
\end{description}

\section{Institutional Trust Dynamics Model} 

In this section we describe how community members' institutional trust is affected by services from organization $\mathcal{H}$. 
In \Cref{sec:rl}, we describe the algorithm that $\mathcal{H}$ uses to learn a policy $\mathcal{P}$ for resource allocation to the community.

\subsection{Network Formation}\label{subsec:network}
In this work, we use simulated networks to represent our communities $G = (V,E)$. We extend the network formation model developed by \citet{christakis2020empirical} to produce realistic seed communities. In brief, individuals are iteratively presented with possible edge-formation decisions, which they choose to take or refuse depending on the utility offered to them. This utility depends on each community member's edge utility function, which includes such factors as: individual attributes, degree, and distance between the possible pairing. These properties fully capture both the attributed and structural qualities of the connection. 
Interested readers can find further details in the original paper \cite{christakis2020empirical}.

\subsection{Citizen Attributes}
\textit{Institutional trust $\tau_v$: } Each community member is assigned a scalar \cite{loi2023much} trust value $\tau_v$ at the beginning of the simulation, where $\tau_v \in [0,1]$ for all $v\in V$. This value represents the degree to which each agent has \textit{institutional trust} in the humanitarian engineering organization $\mathcal{H}$, and is updated as services are provided.

\textit{Service utility $u_v:$ } Recall that each service has some quality $s_v$, which provides community member $v$ with utility $u_v(s_v)$. The value $u_v$ represents the accumulated utility to community member $v$ from all services.

\subsection{The Trust Dynamics Model}

\begin{algorithm}
\caption{Institutional Trust Dynamics Model (ground truth updates) }\label{alg:trust_dynamics}
\KwData{$G = (V,E)$ the network of citizens (and all citizen attributes), $\mathcal{P}$ the policy learned by $\mathcal{H}$, $I$ the number of service iterations}

$\mathbf{s} \gets {\mathcal{P}(G)}$ ; \quad \tcp{get vector of service qualities from learned policy}

\For{$i$ in $0$ ... $I-1$}{

\For{$s_v \in \mathbf{s}$}{
${trust_v} \gets$ True with probability $\tau_v$ \;
\If{$trust_v$}{
    $u_v \gets \frac{i\times u_v + s_v}{i+1}$ ; \quad \tcp{current utility is the average of all services received}
}
}

\For{$v \in V$}{
    $N_v \gets N(v) \cup v$; \quad \tcp{v's neighbors and v}
    $\mathbf{u_N} \gets [u_n \text{ for } n \in N_v]$ ; \quad \tcp{vector of utilities of v's neighbors}
    $fairness \gets 1 - gini(u_N)$; \quad \tcp{Gini coefficient for inequality}
    $influence \gets (\lambda)u_v + (1-\lambda) fairness$; \quad \tcp{positive social influence from services}
    $\tau_v \gets (\delta)influence + (1-\delta)\tau_v$; \quad \tcp{weighted sum of social influence and old trust}
}
}
\Return{$G$}
\end{algorithm}

The trust dynamics model proceeds in 3 stages. 
First, we use policy $\mathcal{P}$ learned by the RL agent to produce a vector \textbf{\textit{s}}, where each element $s_v \in \mathbf{s}$ represents the quality of the service provided to community member $v$, and $\sum_{v\in V}s_v \leq \rho$.
%The RL agent uses the initial network $G$ to learn how to distribute resources $\rho$, where the resources directly translate to service quality. The aim of the RL agent is to allocate resources in such a way that maximizes the organization's utility $\mathcal{U}$.
The organization $\mathcal{H}$ then provides services with the associate service qualities $s_v$, and citizens \textit{only} accept them if they trust the organization. After accepting (or rejecting) services, community members update their utility values accordingly. Here we compute the utility $u_v$ as mean quality of services received up to this point; we choose an average rather than a sum to ensure that values stay within $[0,1]$. 
Once the utilities have been updated, the community members update their trust values as well. The trust update is a linear combination of the utility received from services, the perceived fairness of each citizen's neighborhood, and their prior trust value. The process of providing services and updating trust continues for a set number of service iterations, $I$. \Cref{alg:trust_dynamics} details this process.

Our trust update is motivated by prior research on trust. Though the work of \citet{bhattacharya1998formal} focuses on interpersonal trust, they were one of the first to synthesize a mathematical model of trust that incorporated both economic and psychological factors. We draw inspiration from their model, where actions cause outcomes and therefore consequences, and these provide feedback to the actions. 
However, they do not provide insights regarding the specific factors that might impact trust, so these are motivated by other prior work. In particular, \citet{cabiddu2022users} find: 1) initial trust in an algorithm is a function of a user's propensity to trust, and 2) if a user perceives that an algorithm has positive social influence, there is a higher likelihood of trust building. In addition, \citet{mayer1995integrative, hancock2023and} both emphasize user propensity to trust, as well as reputation of trustee, as important factors when considering overall trust between two individuals. 
% bhattacharya1998formal, mayer1995integrative, hancock2023and, granatyr2015trust
We incorporate these properties into our model; the original trust of each individual $\tau_v$ is assigned at random, we take this to be each individuals' propensity to trust. In addition, we split the ``positive social influence'' observed by \citet{cabiddu2022users} into two components: neighborhood fairness and utility to the individual. 
Work in other sectors has also highlighted the impact of fairness and service quality on institutional trust or trustworthiness \cite{roy2015impact, yeager2017loss, qi2021perceived, lien2014trust, granatyr2015trust}. We use coefficient $\lambda$ to represent the importance of personal utility in the social influence calculation, and $\delta$ to represent the importance of positive social influence in the new trust calculation.
We limit the scope of the fairness observation to each citizen's neighborhood (see line 10 of \Cref{alg:trust_dynamics}), as these are the people each individual would likely communicate with. 
Using these trends found in prior work, we choose specific values for our coefficients $\lambda$ and $\delta$, which we discuss in \Cref{sec:results}.
%We are not aware of any prior work that establishes coefficients (\textit{e.g., } $d$ and $f$ in line 11) for impact of prior trust, fairness, and service quality on current trust. Thus, we choose values for each of these for our work---this can be trivially extended.

\section{Trust-Aware Reinforcement Learning for Resource Allocation} \label{sec:rl}

In this section we describe our trust-aware reinforcement learning algorithm for resource allocation. The outputted policy is used to produce a vector $\mathbf{s}$ in \Cref{alg:trust_dynamics} in the previous section.

\subsection{Deep Deterministic Policy Gradient (DDPG)} \label{subsec:ddpg}
DDPG \cite{lillicrap2015continuous} is a model-free, off-policy, actor-critic method ideal for use in complex environments, particularly those with continuous action spaces. In brief, DDPG employs two neural networks: the actor, which generates actions, and the critic, which evaluates these actions through a Q-function \cite{sumiea2024deep}. Due to its ability to learn complex policies in dynamic environments, DDPG is ideal for learning resource allocations on networks \cite{sumiea2024deep}.
The focus of this paper is not on the detailed description or development of the RL architecture, but rather on leveraging DDPG in a simple problem setting to produce insights on trust. 
Those interested in technical DDPG details can read further \cite{lillicrap2015continuous, sumiea2024deep, tan2021reinforcement}.

Below we give details regarding the RL algorithm's problem formulation. In the next subsection we discuss the simulated environment the RL agent interacts with. Note that answering \textbf{RQ1} and \textbf{RQ2} require slightly different state spaces and simulated environments. We describe these differences here and give further details in \Cref{tab:models} and \Cref{sec:results}.

\textbf{State space: }
The state space is a tuple which requires a representation of the following essential components:
1)~\textit{Social network:} Our chosen representation of the social network is the adjacency matrix for $G$; it is important that the RL agent is aware of the edges that exist on the network so it can correctly predict trust dynamics.
2)~\textit{Service utility:} The RL agent keeps track of the accumulated utility from services to each community member. These values are initialized to $0$ and is represented by a vector of length $|V|$, which we call $\mathbf{u}$. 
3)~\textit{Institutional trust:} The RL agent must keep track of the institutional trust of each community member. When answering \textbf{RQ1} (the trust-aware model), we represent trust by a vector of length $|V|$, where the RL agent has access to the initial trust value $\tau_v$ of each citizen $v$. When answering \textbf{RQ2} (the learned-trust model), the RL agent does not have access to these trust values. Instead, it must learn these values via Bayesian updates. The representation of trust for each community member is a pair of values $\alpha$ and $\beta$ which serve as the parameters for a Beta distribution, \textit{i.e., } Beta$(\alpha, \beta)$. These values are updated when the agent observes a community member accepting or rejecting services (see \Cref{alg:trust_dynamics_sim_rq2}). Then, the representation of trust in the learned-trust model is a vector of tuples $(\alpha_v, \beta_v)$ of length $|V|$.

\begin{table}[h]
\centering
    \caption{Our model variants and the information scope for each.}
    \label{tab:models}
    \begin{tabularx}{\linewidth}{r|p{2.7in}|p{2in}}
        % \toprule
    \textsc{The Model} & \textsc{Description} & \textsc{Information Scope}\\ 
    \midrule
    Trust-unaware & As a baseline, the RL agent learns a resource allocation without any knowledge of trust. & Network $G$\\
    Trust-aware & To answer \textbf{RQ1}, the RL agent learns a resource allocation with knowledge of initial trust values. & Network $G$, $\tau_v$ for all $v \in V$, whether $v$ accepts service $s_v$, and trust update rule\\
    Learned-trust & To answer \textbf{RQ2}, the RL agent learns a resource allocation while also learning distributions for each community members' trust value.& Network $G$, whether $v$ accepts service $s_v$, and trust update rule
        
\end{tabularx}
\end{table}

\textbf{Action space: }
The organization $\mathcal{H}$ has a budget of $\rho$. As described in \Cref{sec:ps}, each iteration, the humanitarian engineering organization $\mathcal{H}$ provide services of quality $\mathbf{s}$. Each action $s_v \in \mathbf{s}$ is restricted such that $\sum_{v\ \in V} s_v \leq \rho$, and each $s_v \in [0,1]$. The goal of the RL agent is to learn the vector $\mathbf{s}$ that will maximize the reward function. 

\textbf{Reward function: }
As described in \Cref{sec:ps}, the organization $\mathcal{H}$ has utility metric $\mathcal{U}$. We stated that this metric may be a function of how many people are affected by services, the quality of the services, any remaining resources from $\rho$, and the private attribute $c$ representing the willingness of the organization $\mathcal{H}$ to reduce the quality of services to stay under budget. 
In this work, given a utility vector $\mathbf{u}$ with elements $u_v$, action vector $\mathbf{s}$ with elements $s_v$, attribute $c$, and resource constraint $\rho$, our reward function (or organization utility metric) is defined as follows: 
$$ \mathcal{U}(u,S, c,\rho) = (1 - c)\left(\frac{\sum_{v \in }u_v}{2|V|} + \frac{|u|_{\gamma}}{2|V|}\right) + c\left(\rho-\sum_{v\in V}s_v\right)
$$
where $|u|_{\gamma} = \sum_{u_v \in \mathbf{u}} \mathbbm{1}_{u_v \geq \gamma}$ \textit{i.e.,} the number of elements over some $\gamma$-threshold.
Recall that while learning a policy, the RL agent is making \textit{predictions} regarding utility vector $\mathbf{u}$. 
%Once all services have been provided, the RL agent knows the true value of $\mathbf{u}$, which it can learn from surveys, for example.
The first term of $\mathcal{U}$ gives us the average of mean service quality over all citizens and the number of citizens affected by the service, and we weight this term by $(1-c)$. Here, we make an assumption that the organization $\mathcal{H}$ cares equally about helping many people and providing high quality services. The second term calculates remaining resources after the action has been taken, and we weight this by $c$ to represent the importance the RL agent will place on this term. Note that since all relevant values are in $[0,1]$, our organization's utility $\mathcal{U}$, and thus the RL agent rewards, will also be in $[0,1]$. 
Our reward function is designed to capture a wide range of behaviors when $c$ is varied, and we show these results in \Cref{sec:results}.

\subsection{Simulated Environments}
\begin{algorithm}[h!]
\caption{Simulated Environment for Trust Dynamics (Trust-aware: RQ1)}\label{alg:trust_dynamics_sim_rq1}
\KwData{$G = (V,E)$ the network of citizens (and initial citizen trust $\tau_v$), $\mathbf{s}$ the learned service vector, $I$ the number of steps per episode}

\For{$i$ in $0$ ... $I-1$}{

\For{$s_v \in \mathbf{s}$}{
${trust_v} \gets$ True with probability $\tau_v$ \;
\If{$trust_v$}{
    $u_v \gets s_v$ ; \quad \tcp{org observes citizen accepting/rejecting service, does not calc avg}
}
}

\For{$v \in V$}{
    $N_v \gets N(v) \cup v$; \quad \tcp{v's neighbors and v}
    $\mathbf{u_N} \gets [u_n \text{ for } n \in N_v]$ ; \quad \tcp{vector of utilities of v's neighbors}
    $fairness \gets 1 - gini(u_N)$; \quad \tcp{Gini coefficient for inequality}
    $influence \gets (\lambda)u_v + (1-\lambda) fairness$; \quad \tcp{positive social influence from services}
    $\tau_v \gets (\delta)influence + (1-\delta)\tau_v$; \quad \tcp{weighted sum of social influence and old trust}
}
}
\end{algorithm}
\begin{algorithm}[h!]
\caption{Simulated Environment for Trust Dynamics (Learned-trust: RQ2)}\label{alg:trust_dynamics_sim_rq2}
\KwData{$G = (V,E)$ the network of citizens (including $\tau_v$, unobservable by agent), $\mathbf{s}$ the learned service vector, $I$ the number of steps per episode}

\For{$v \in V$}{
$[\alpha_v,\beta_v] \gets [1,1]$\;
}
\For{$i$ in $0$ ... $I-1$}{

\For{$s_v \in \mathbf{s}$}{
${trust_v} \gets$ True with probability $\tau_v$ \;
\If{$trust_v$}{
    $u_v \gets s_v$ ; \quad \tcp{org observes citizen accepting/rejecting service, does not calc avg}
    $[\alpha_v, \beta_v] \gets [\alpha_v + 1, \beta_v]$ ; \quad \tcp{update success prob}
}
\Else{
$[\alpha_v, \beta_v] \gets [\alpha_v, \beta_v +1]$ ; \quad \tcp{update failure prob}
}
}
\For{$v \in V$}{
    $\widehat{\tau_v} \gets \frac{\alpha_v}{\alpha_v+\beta_v}$ ; \quad \tcp{agent uses mean of distribution}
    $N_v \gets N(v) \cup v$; \quad \tcp{v's neighbors and v}
    $\mathbf{u_N} \gets [u_n \text{ for } n \in N_v]$ ; \quad \tcp{vector of utilities of v's neighbors}
    $fairness \gets 1 - gini(u_N)$; \quad \tcp{Gini coefficient for inequality}
    $influence \gets (\lambda)u_v + (1-\lambda) fairness$; \quad \tcp{positive social influence from services}
    $\tau_v \gets (\delta)influence + (1-\delta)\tau_v$; \quad \tcp{update true trust, but agent cannot observe this}
    $\widehat{\tau_v} \gets (\delta)influence + (1-\delta)\widehat{\tau_v}$; \quad \tcp{update trust prediction}
    $\alpha_v \gets \alpha_v + 0.5(\widehat{\tau_v})$ ; \quad \tcp{update success prob according to confidence in trust prediction}
    $\beta_v \gets \beta_v + 0.5(\widehat{\tau_v})$ ; \quad \tcp{update failure prob according to confidence in trust prediction}
    
}
}
\end{algorithm}
DDPG is model-free, meaning it does not learn a model for its environment; we provide a simulated environment for the RL agent to interact with. We design the simulated environment to realistically model the understanding that $\mathcal{H}$ may have of the community it serves, \textit{i.e.,} it closely follows \Cref{alg:trust_dynamics}, with some small differences. First, the organization must approximate the utility agent $v$ receives from a service with quality $s_v$ (line 5, \Cref{alg:trust_dynamics_sim_rq1} and line 7, \Cref{alg:trust_dynamics_sim_rq2}). 
For the learned-trust model, where the agent cannot observe the consumer trust $\tau_v$, we have the RL agent observe the coin flip (\textit{i.e.,} whether or not the individual accepts the service) rather than the trust value itself (line 5, \Cref{alg:trust_dynamics_sim_rq2}). The RL agent then uses this value to update the posterior distribution for each individual.
\Cref{alg:trust_dynamics_sim_rq1} below shows the simulated dynamics of the environment that the trust-aware RL agent interacts with, and \Cref{alg:trust_dynamics_sim_rq2} shows the same for the learned-trust agent. The for-loop over iterations represents the steps taken each episode.

\section{Simulation Results} \label{sec:results}
In this section we present the results of policy adoption simulations from our trained RL model. As a reminder, we begin by initializing our social network, and assigning each community member $v \in V$ a random trust value $\tau_v$ drawn from a distribution. With this data, the humanitarian engineering organization $\mathcal{H}$ learns a policy that dictates how resources should be distributed in the community (\textit{i.e.,} vector $\mathbf{s}$). Then, we simulate the distribution of resources, and analyze several metrics to determine the organization's success, as well as the impact on the community members.

Before presenting results, we describe our simulation parameters. We consider a network of 15 nodes, each representing one household in a community. We keep our numbers small for simplicity and interpretability of the action and state space. The size of the network could be extended with some hyperparameter tuning. 
As we described in \Cref{subsec:network}, this social network is produced using the network formation process developed in \cite{christakis2020empirical}, and we show examples of these networks later in the section (see \Cref{fig:rq1_ntwks}) when we present our results. We simulate 25 service iterations, \textit{i.e.,} the number of rounds the organization is able to provide a service to the community---$I$ in our algorithms. 
We set the resource constraint to $\rho = 1$; this can be easily extended as the sum of services will scale to the total resource constraint. 

In the following subsections, we vary two parameters: $c$, the willingness of $\mathcal{H}$ to reduce the quality of services to conserve resources, and $\tau$, the distribution of initial institutional trust values across community members. We consider $c \in \{0,0.25,0.5,0.75,1.0\}$, and $\tau \in \{\text{Beta}(2,8), \text{Beta}(2,6), \text{Beta}(2,4), \text{Beta}(2,2), \text{Beta}(4,2), \text{Beta}(6,2), \text{Beta}(8,2)\}$. 
We choose Beta distributions to represent trust as these are appropriate for random distributions of proportions. When we present results, we use the mean of the distribution to refer to the trust level of the population, \textit{i.e.,} $\overline{\tau}\in \{ 0.20, 0.25, 0.33, 0.50, 0.67, 0.75, 0.80\}$. We choose $\lambda = 0.8$ and $\delta = 0.5$ (from Algorithms \ref{alg:trust_dynamics}, \ref{alg:trust_dynamics_sim_rq1}, and \ref{alg:trust_dynamics_sim_rq2}); we equally weight social influence and prior trust, and emphasize individual utility over fairness. These values could be changed quite easily in future work. 
We run each simulation five times, producing a new network, training a new policy, and simulating the services from the policy on the network. We present the average of those iterations for both research questions. 
 
\subsection{RQ1: Trust-Aware Policy}
In \textbf{RQ1} we assume the institutional trust value of each community member, $\tau_v$, is public. We ask: Can the humanitarian engineering organization $\mathcal{H}$ learn the optimal policy to allocate technology while minimally eroding trust? How effective is this policy when compared to one that does not incorporate trust? Recall that our trust-aware reinforcement learning algorithm accounts for trust in the state space, and approximates the service utility of each community member $u_v$ according to the current trust value $\tau_v$. 
The RL agent does not maximize trust explicitly in the reward function (see \Cref{subsec:ddpg}), as this value itself is not useful to the organization $\mathcal{H}$. 
Rather, we assume that the use in minimally eroding trust comes from the downstream ability to, with high likelihood, increase accumulated service utility of community members due to the increased probability of accepting services. The service utility is accounted for in the organization's utility function $\mathcal{H}$, and in this way encapsulates trust. We compare this algorithm a trust-unaware agent, which is completely unaware of the trust dynamics and thus does not update service utility with respect to trust.

\begin{figure}[t]
    \centering
    \begin{tabular}{@{}c c c@{}}
    \includegraphics[width = .32\linewidth]{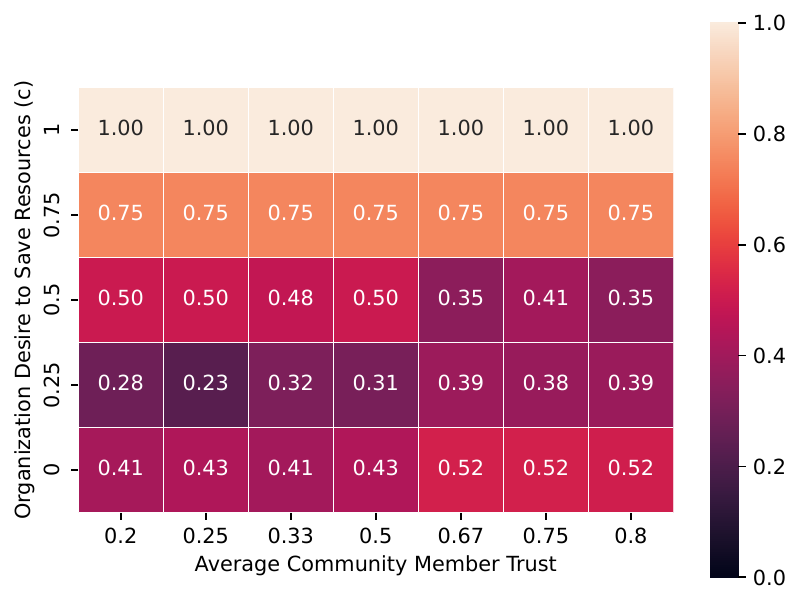} &  
    \includegraphics[width = .32\linewidth]{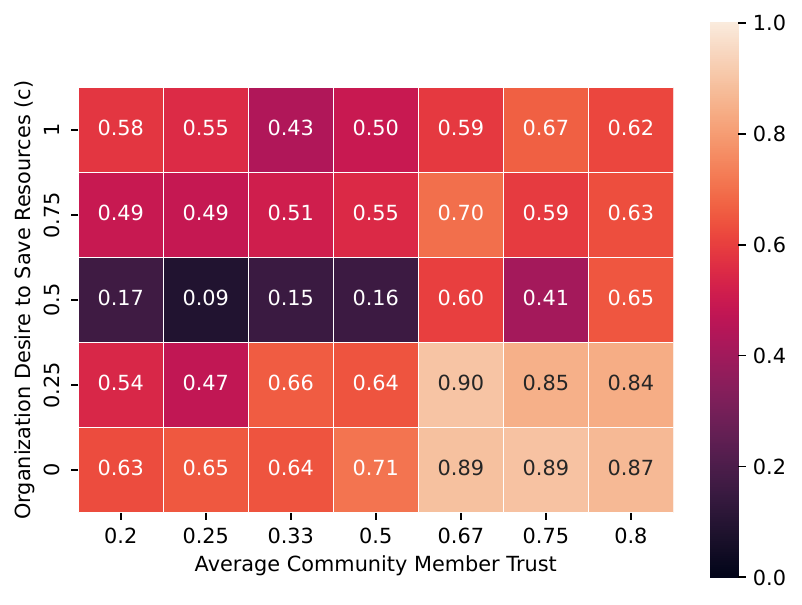} & \includegraphics[width = .32\linewidth]{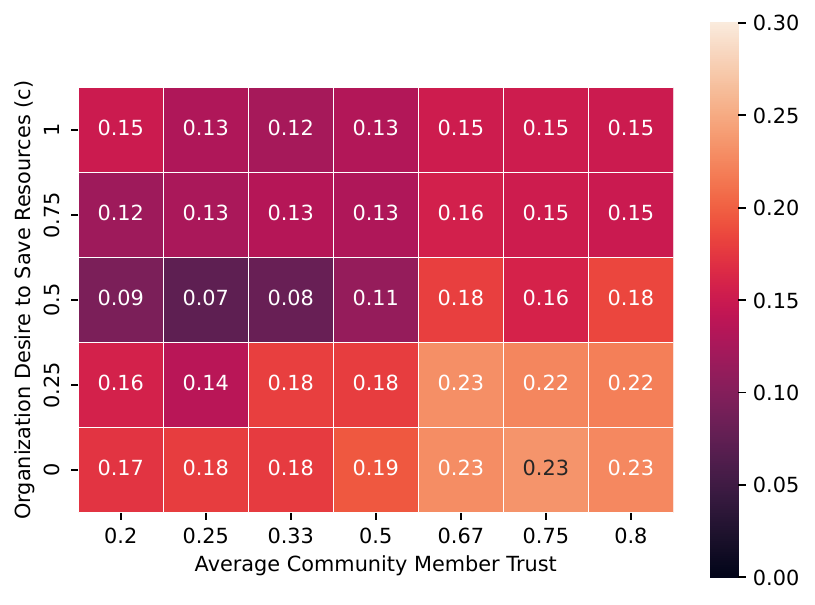}\\
    \small{(a) Trust-aware $\mathcal{U}$} & \small{(b) Trust-aware fairness}  & \small{(c) Trust-aware average trust}\\
    \includegraphics[width = .32\linewidth]{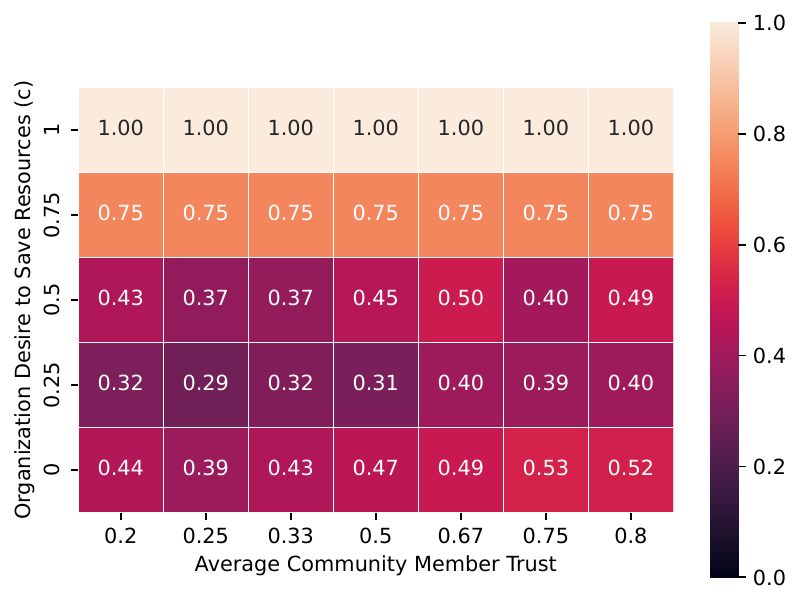} &  
    \includegraphics[width = .32\linewidth]{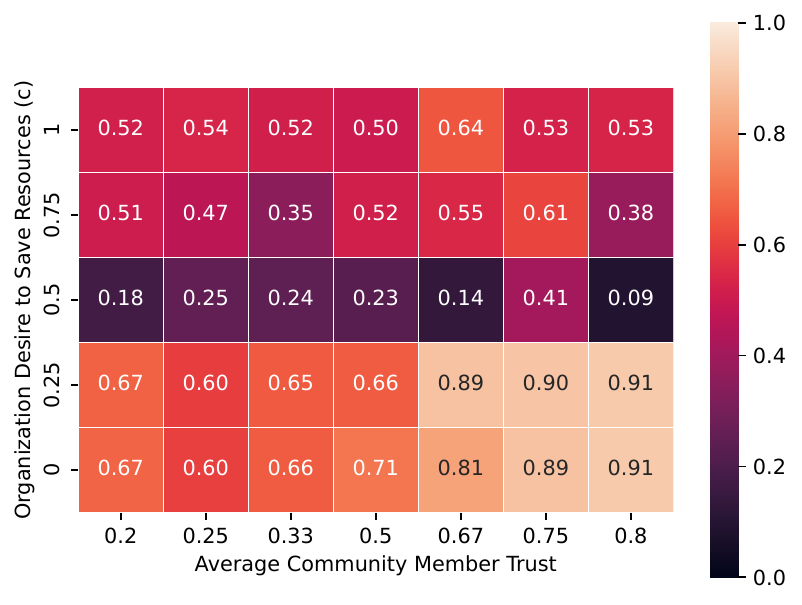} & \includegraphics[width = .32\linewidth]{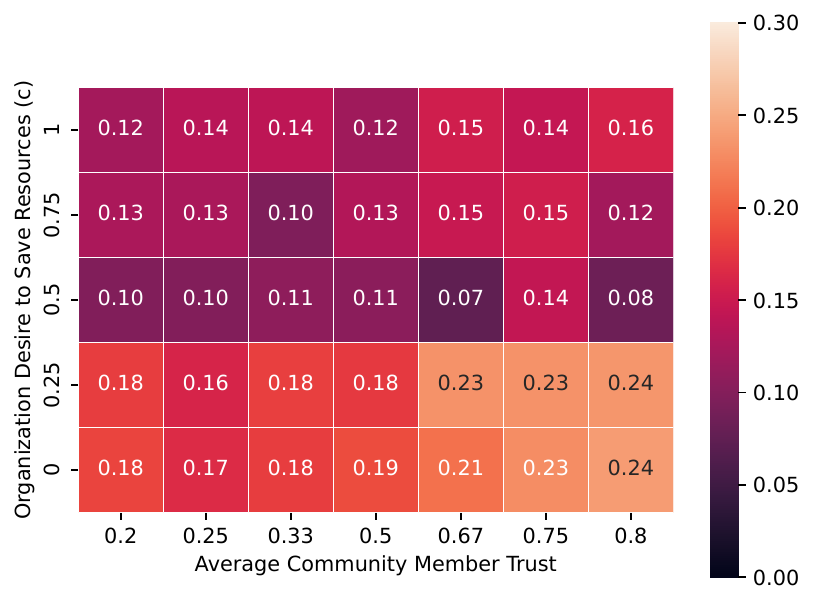}\\
    \small{(d) Trust-unaware $\mathcal{U}$} & \small{(e) Trust-unaware fairness}  & \small{(f) Trust-unaware average trust}
    \end{tabular}
    \caption{Organization success (or utility $\mathcal{U}$) (a \& d), global fairness of resource allocation (b \& e), and average individual trust (c \& f) for our trust-aware and unaware RL algorithms. The trends for each algorithm are quite similar---\textbf{high $c$ values, or high desire to save resources, lead to success for the RL agent, but lost trust.} For low $c$ values, higher initial average trust allows the RL agent to learn a more successful policy that positively impacts the community, maximizing fairness as well as trust. Notice that the scale for average trust is different than the other metrics (see color bar).}
    \label{fig:rq1_ind}
\end{figure}
\begin{figure}[h]
    \centering
    \begin{tabular}{@{}c c c@{}}
    \includegraphics[width = .32\linewidth]{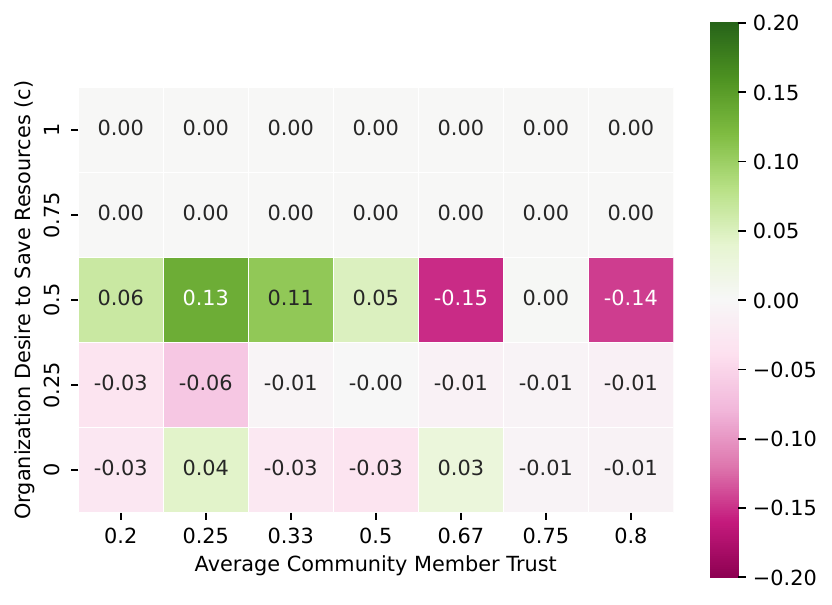} &  
    \includegraphics[width = .32\linewidth]{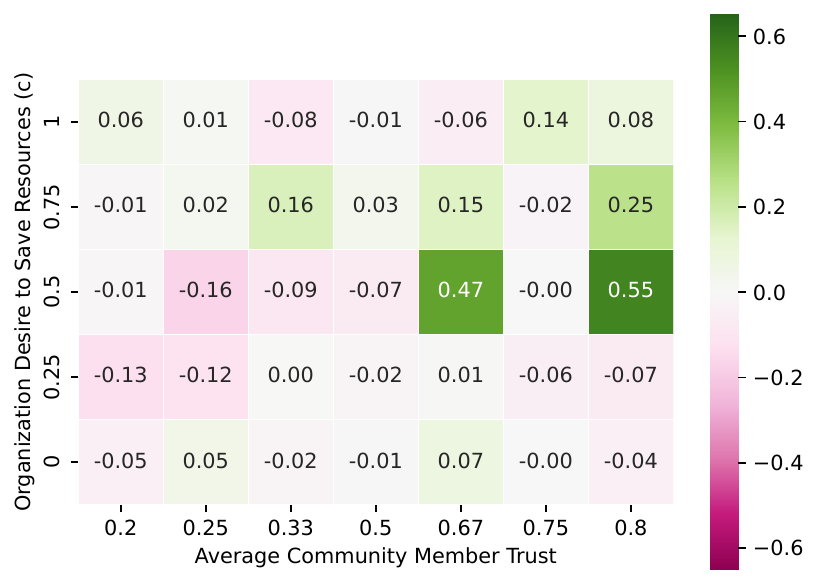} & \includegraphics[width = .32\linewidth]{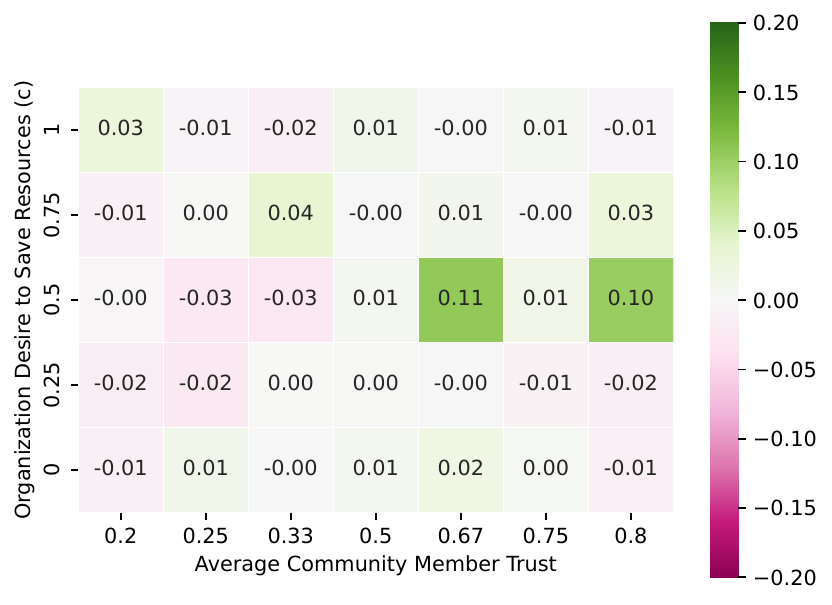}\\
    \small{(a) Trust-aware $\mathcal{U}$ diff} & \small{(b) Trust-aware fairness diff}  & \small{(c) Trust-aware average trust diff}
    \end{tabular}
    \caption{Raw difference for each metric between trust-aware and trust-unaware RL algorithms. Green cells represent the trust-aware algorithm outperforming trust-unaware, pink is the opposite, and white is neutral. 
    \textbf{Our trust-aware algorithm most outperforms unaware in fairness and average trust when $c = 0.5$ and $\overline{\tau} \geq 0.67$. However, this boost in fairness and trust comes at the cost of organization success.} For other $c$ values, we find that the additional certainty does not help performance, or has inconsistent effects.}
    \label{fig:rq1_comp}
\end{figure}

\begin{figure}[b]
    \centering
    \begin{tabular}{@{}c c c@{}}
    \includegraphics[width = .27\linewidth, trim={3cm 3cm 3cm 3cm}, clip]{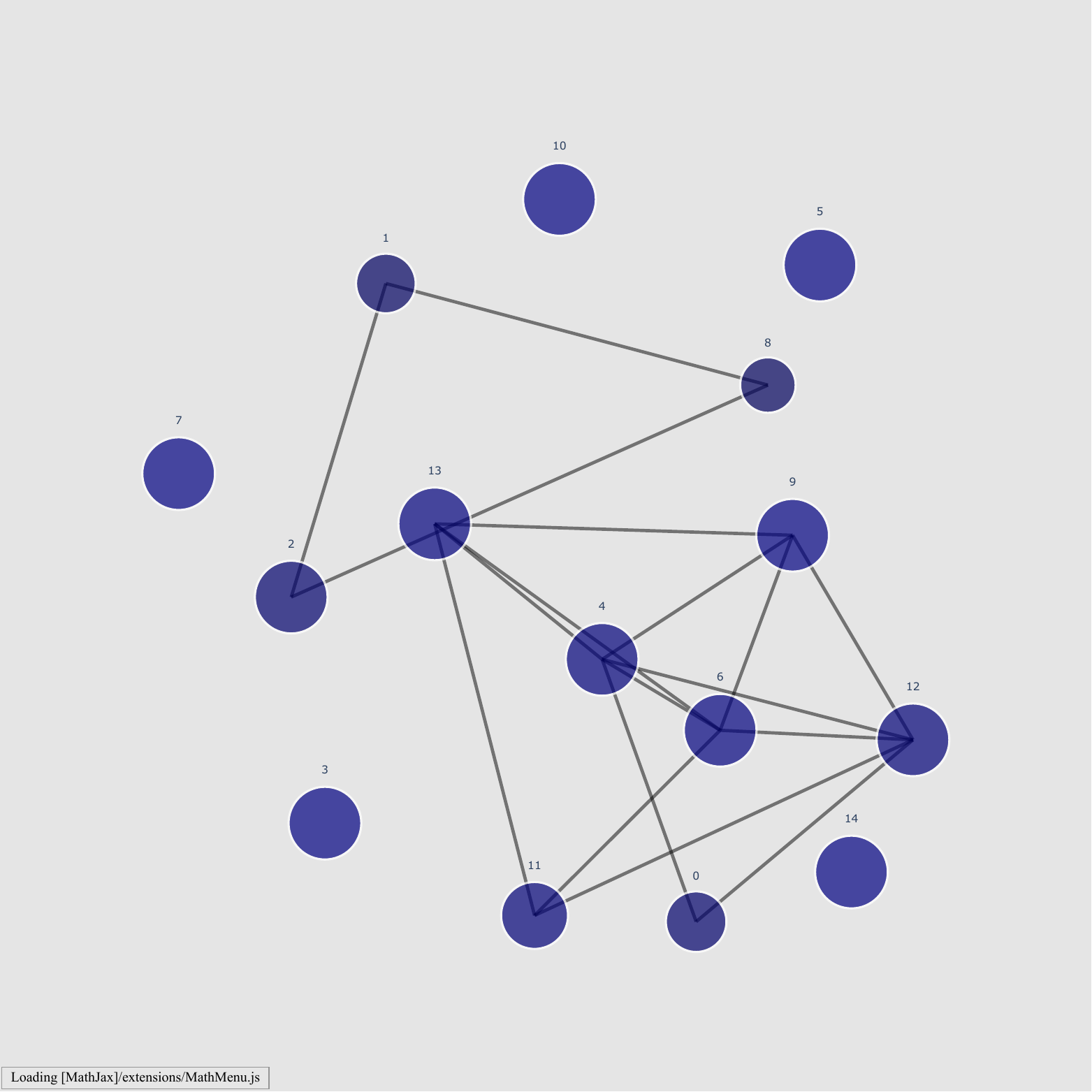} &  
    \includegraphics[width = .27\linewidth, trim={3cm 3cm 3cm 3cm}, clip]{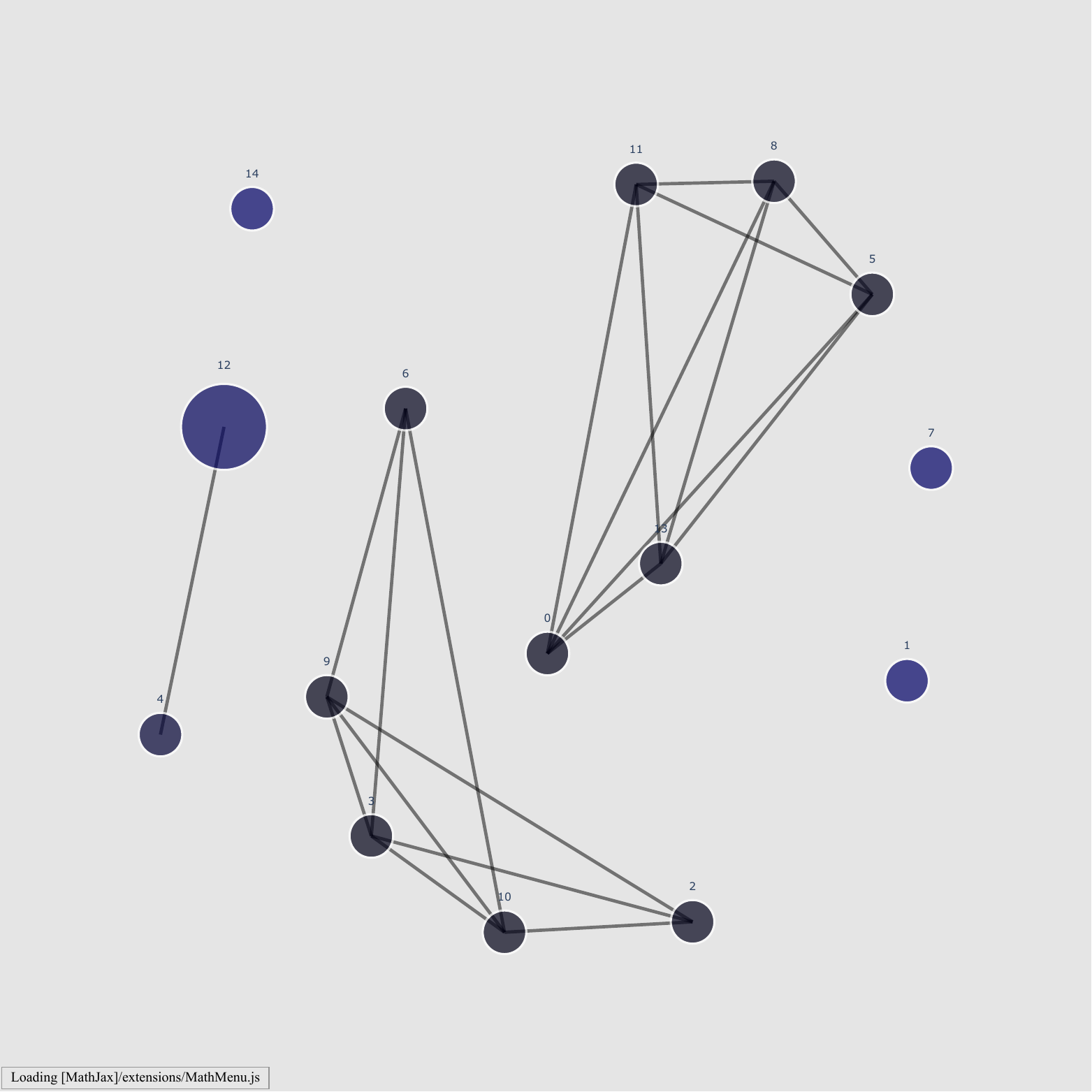} & \includegraphics[width = .27\linewidth, trim={3cm 3cm 3cm 3cm}, clip]{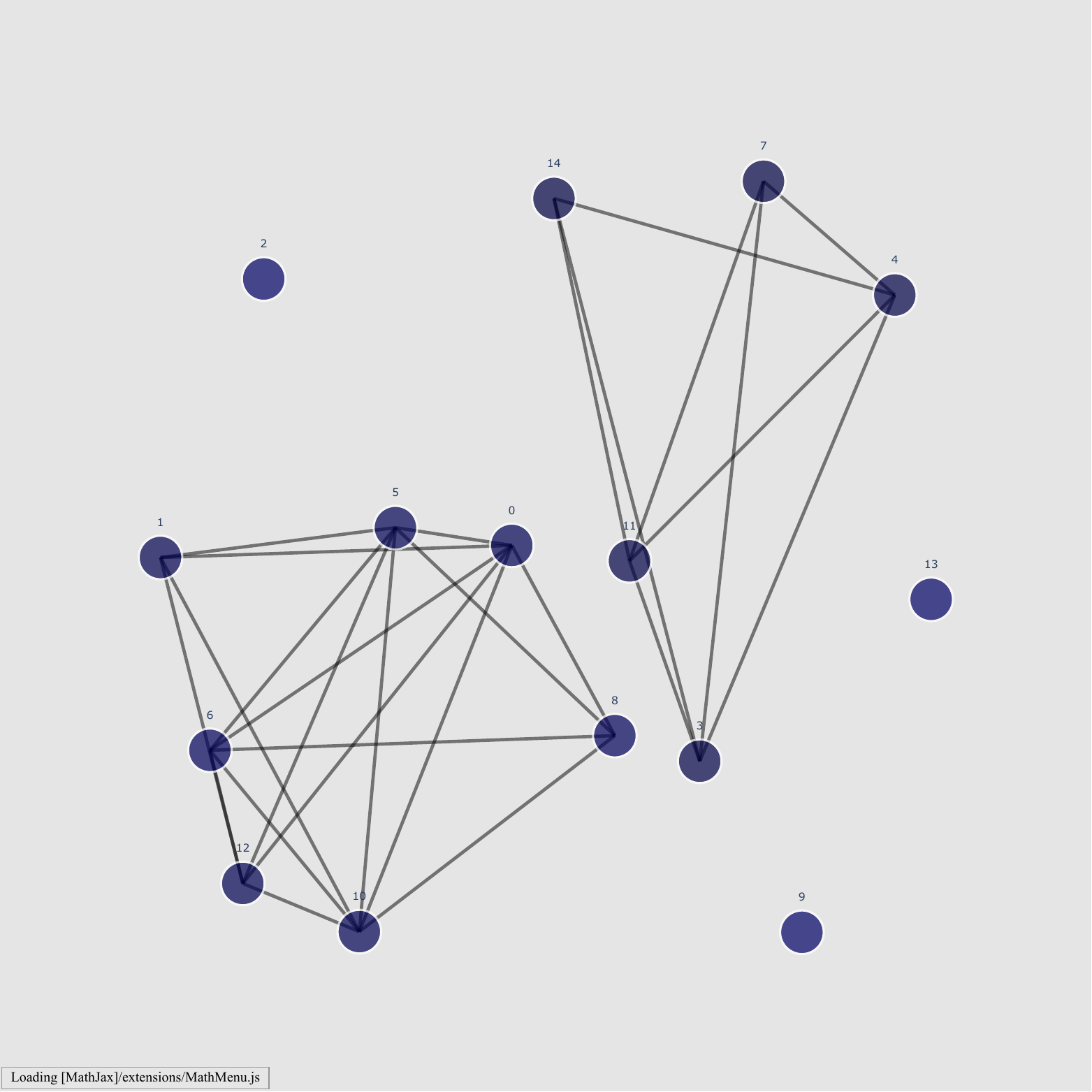}\\
    \small{(a) Trust-aware model, $c=0.0, \overline{\tau} = 0.8$} & \small{(b) Trust-aware model, $c=0.50, \overline{\tau} = 0.8$}  & \small{(c) Trust-aware model, $c=1.0, \overline{\tau} = 0.8$}
    \end{tabular}
    \caption{We show social networks for select iterations of the trust-aware algorithm after services have been provided. We fix $\overline{\tau} = 0.8$ for ease of comparison. The color of node $v$ indicates the updated $\tau_v$ value after services have been provided; bluer nodes have higher values. The size indicates the accumulated utility $u_v$; larger nodes have accepted more services. For high values of $c$, community members do not receive any services; as $c$ decreases, service values increase (a). \textbf{Isolated nodes tend to be more trusting, as they always perceive perfect fairness in their neighborhoods.}}
    \label{fig:rq1_ntwks}
\end{figure}

To evaluate effectiveness of each of these algorithms, we investigate multiple metrics, including the organization's utility $\mathcal{U}$, the global fairness in distribution of utilities $u_v$ of each community member $v$, and average trust $\tau_v$ of each community member $v$. We determine the fairness of the utility distribution using $1-gini(u)$ \cite{dorfman1979formula}. In \Cref{fig:rq1_ind} we show results from our simulations; trust-aware in the top row, trust-unaware in the bottom row. The trust-aware and unaware algorithms follow similar trends for each metric; we discuss these trends before moving on to the differences between the models. When $c$, or organization $\mathcal{H}$'s desire to conserve resources by degrading service quality, is high ($c \geq 0.75$), it is simpler for the RL agent to find a resource allocation that maximizes the reward function (a \& d)---this involves allocating services with 0 or very low quality to all community members. Using our fairness metric, $1-gini(u)$, this is considered a somewhat ``fair'' allocation (b \& e). However, trust suffers (c \& f), as the utility of the service being provided to each of them is minimal. Note also that as initial trust increases, so does fairness and final average trust of community members. Since the resource allocation is not \textit{exactly} $0$, low trust leads to rejection of services and thus less fair outcomes. See \Cref{fig:rq1_ntwks}c for an example network where $c=1$ and initial trust is high ($\overline{\tau} = 0.8$). 

When $c = 0.5$, the agent has some difficulty balancing the objectives of saving resources and providing services. To maximize the utility function $\mathcal{U}$, the RL agent again learns to provide services with approximately 0 quality to most nodes per iteration, but provides a very low quality service to select nodes. We see these small services being provided to select nodes because of our term in the utility function $\mathcal{U}$ that rewards the RL agent for providing any approximately non-zero quality service to a node. 
See \Cref{fig:rq1_ntwks}b as an example; most nodes have accumulated 0 service utility, but node 12 has accumulated much more. While this helps the agent achieve its goal, it also decreases the fairness in the community, and thus decreases trust as well.  

For low values of $c$, \textit{i.e.,} when $c \leq 0.25$, we see a marked increase in the organization's success as average trust moves from $0.50$ to $0.67$. In higher trust scenarios, the RL agent is able to choose a more successful policy, as community members accept services maximizing fairness and average community member trust. 
%We do note that for many parameter values, average trust actually decreases as services are provided (see \Cref{fig:rq1_ind}c \& f). 
However, even in low trust scenarios, ($\overline{\tau} = 0.25$), the RL agent is able to learn a policy that provides services to many community members. See \Cref{fig:rq1_ntwks}a for an example of a community in a high trust scenario ($\overline{\tau} = 0.8$).

While both models exhibit similar general trends for each metric, to answer \textbf{RQ1}, we are interested in understanding under which scenarios our trust-aware reinforcement learning algorithm is more \textit{successful} than trust-unaware. We show the raw difference in each metric between the models in \Cref{fig:rq1_comp} (\textit{i.e.,} the second row of \Cref{fig:rq1_ind} subtracted from the first row). \Cref{fig:rq1_comp}a shows the raw difference in the organization's success, or the utility function $\mathcal{U}$. By the presence and magnitude of green cells, we can see that our trust-aware algorithm is more \textit{successful} than trust-unaware when $c$, or the organization's willingness to decrease quality to save resources, is $0.5$, and $\overline{\tau}$, or average community member trust, is less than $0.67$.
While the RL agent has some difficulty learning successful policies when $c = 0.5$ due to uncertainty in the utility function $\mathcal{U}$, additional information in the trust-aware simulation is able to rectify this somewhat.
The improved success, however, does not translate to increased fairness or average trust (\Cref{fig:rq1_comp}b and c). We hypothesize that the RL agent uses the additional trust information to forgo distributing resources to low-trust agents, which in turn exacerbates issues of fairness and trust in the community. 
However, when trust values are high, the organization does not succeed. Here, we conversely hypothesize that the RL agent uses the additional trust information to allocate higher quality resources to all agents, leading to increased trust and fairness, but decreased success of the organization as it is still aiming to conserve resources. 
When $c$ values are high or low, it is difficult to discern any pattern regarding optimality of either model. We hypothesize that some of these fluctuations may be coming from differing network structures, which we discuss in \Cref{sec:disc}.
%In general, we see that high trust values lead to success for the trust-aware model when $c = 0$, and higher fairness when $c=0, 0.75,$ and $1$, but the trust-unaware model dominates for $c=0.25$. 
These results provide us with a nuanced answer to our research question: the humanitarian engineering organization $\mathcal{H}$ can learn successful policies 
using trust-aware RL algorithms, but this may come at the cost of community fairness and trust, especially when $c=0.5$, and initial trust is low. 
Since our trust-aware model does not dominate trust-unaware in all scenarios, we consider a different representation of trust in \textbf{RQ2}.

\begin{figure}
    \centering
    \begin{tabular}{@{}c c c@{}}
    \includegraphics[width = .32\linewidth]{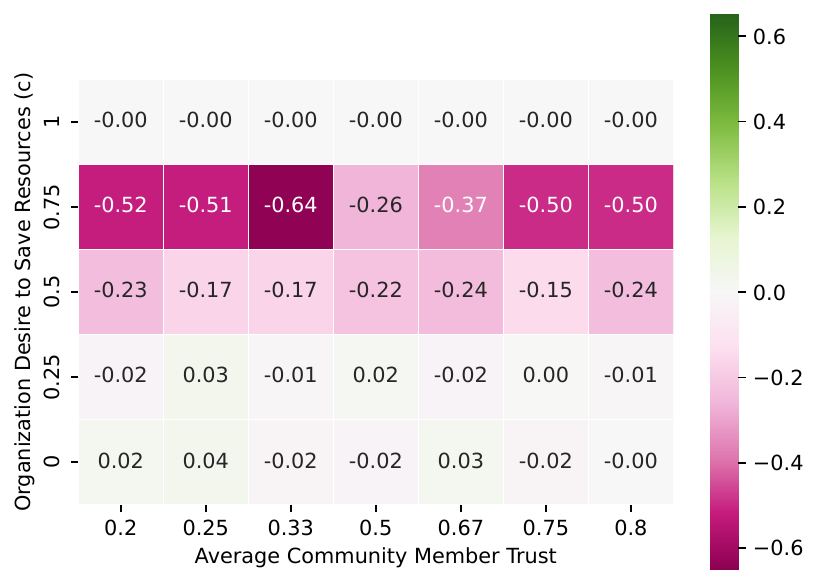} &  
    \includegraphics[width = .32\linewidth]{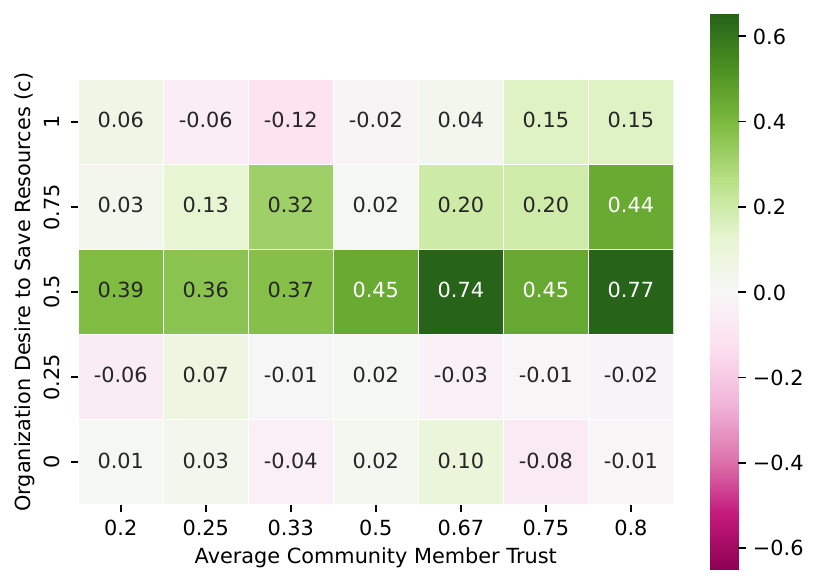} & \includegraphics[width = .32\linewidth]{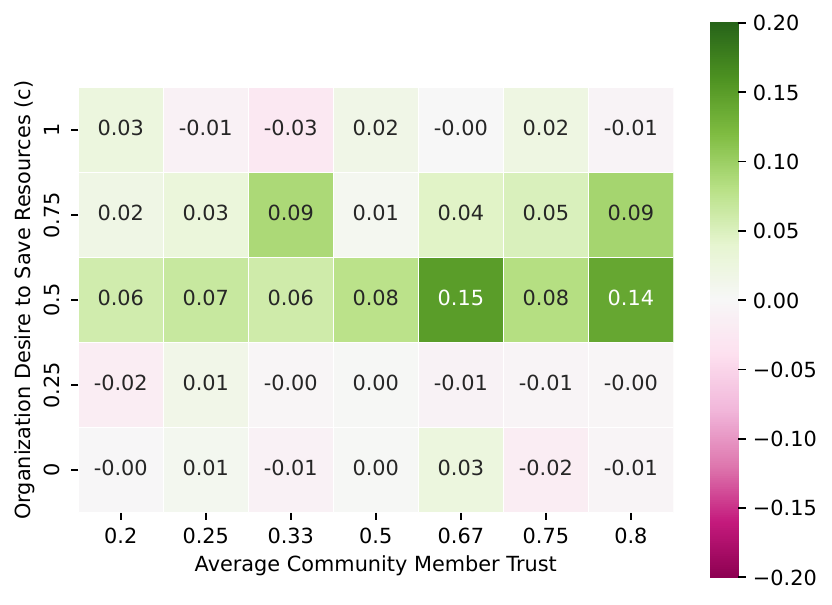}\\
    \small{(a) Learned-trust $\mathcal{U}$ diff} & \small{(b) Learned-trust fairness diff}  & \small{(c) Learned-trust average trust diff}
    \end{tabular}
    \caption{Raw difference for each metric between learned-trust and trust-unaware RL algorithms. Green cells represent the learned-trust algorithm outperforming trust-unaware, pink is the opposite, and white is neutral. Notably, the learned-trust model does not produce a sharp dip in fairness (and therefore trust) at $c=0.5$; this is due to the more equitable resource distributions learned by the agent. \textbf{In most cases, the learned-trust model dominates trust-unaware for fairness and trust, but organization success suffers.}}
    \label{fig:rq2_ind}
\end{figure}
   
\subsection{RQ2: Learned-Trust Policy}
In \textbf{RQ2} we assume instead that the trust values $\tau_v$ are private. We ask: Can $\mathcal{H}$ learn these trust values while learning the policy? How does this lack of knowledge affect the success of the policy? Recall that to answer \textbf{RQ2}, the RL agent tracks parameters that describe a distribution that predicts the trust of each community member, rather than the trust value itself (see \Cref{alg:trust_dynamics_sim_rq2} for details). In summary, the RL agent begins with the prior assumption that all community members have their trust value $\tau_v$ drawn from $\text{Beta}(1,1)$. As the RL agent gathers more information (by observing acceptances and rejections of services by community members), it updates its prediction regarding the trust value of each individual. It then takes the mean of this distribution as the predicted trust value. 

We calculate the same metrics as we did in \textbf{RQ1} to evaluate the success of the policy learned by this RL agent (\Cref{fig:rq2_ind}). At first glance, it is clear that the learned-trust model produces entirely different results from the trust-unaware model, and trust-aware as well.
Most notably, the learned-trust model is able to better overcome the sharp dip in fairness and trust at $c=0.5$ and $c=0.75$ (\Cref{fig:rq2_ind}b and c), due to more equitable distribution of resources. Specifically, we avoid this dip because the RL agent does not learn policies in which only one or two community members are receiving services. However, as we noted earlier, in turn organization success suffers (\Cref{fig:rq2_ind}a). These results indicate that a more conservative Bayesian estimate of trust leads to fairness and trust-increasing policies, especially when the organization's desire to save resources is high. Because this RL agent avoids any extreme trust values, we hypothesize that this encourages less extreme policies, which are more fair and trust-improving, even without the knowledge of original community members' trust values. 
Thus, to answer our second research question---yes, the organization $\mathcal{H}$ can learn trust values using other signals given by community members. In fact, those signals lead to more conservative predictions by the RL agent, which in some cases improve the welfare of community members, but not the success of the organization.

\section{Discussion} \label{sec:disc}

\subsection{Quota System Intervention }\label{sec:quota}
While answering \textbf{RQ1} and \textbf{RQ2}, we considered scenarios where the humanitarian engineering organization $\mathcal{H}$ often prioritized saving resources over helping community members $\textit{i.e.,}$ when attribute $c$ had a high value.
While the organization was able to ``successfully'' distribute resources under these conditions (\textit{i.e.,} maximize its own utility metric $\mathcal{U}$), community members suffered---both in fairness and institutional trust. Our learned-trust model was able to counteract this somewhat. However, to ensure further that community members are not harmed by RL policies, we design a simple intervention and explore it briefly. We implement a quota system---the organization $\mathcal{H}$ is required by some higher governmental entity to serve at least $k$-many community members each time step, else their utility $\mathcal{U}$ will decrease. Specifically, when $|u|_{\gamma} < \frac{|V|}{2}$, $\mathcal{U} = \mathcal{U} - \frac{1}{2}(1-\frac{|u|_{\gamma}}{|V|})$, where $|u|_{\gamma}$ is the number of elements in the utility vector $\mathbf{u}$ over some $\gamma$ threshold. In other words, when $\mathcal{H}$ provides service utility to less than half the community members, its utility $\mathcal{U}$ will decrease proportionally with the number of community members who are receiving approximately $0$ utility from services.  

\begin{figure}[b]
    \centering
    \begin{tabular}{@{}c c c@{}}
    \includegraphics[width = .32\linewidth]{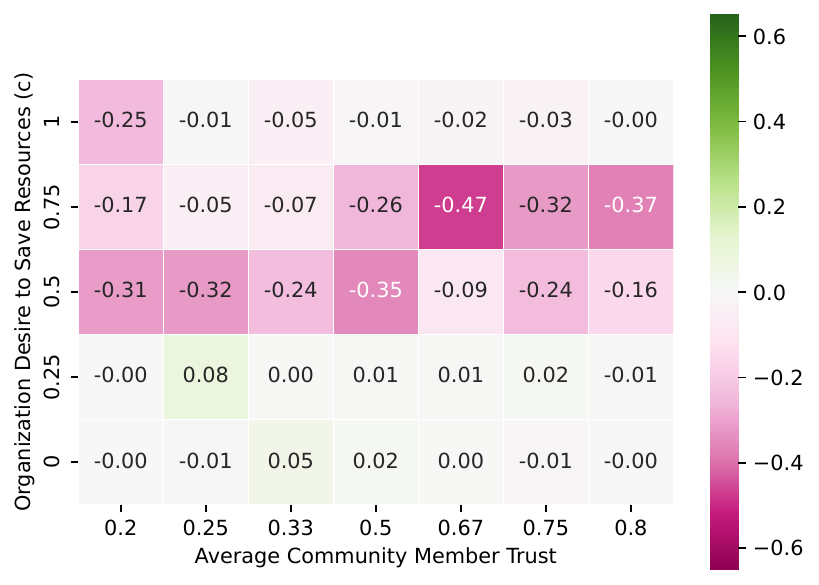} &  
    \includegraphics[width = .32\linewidth]{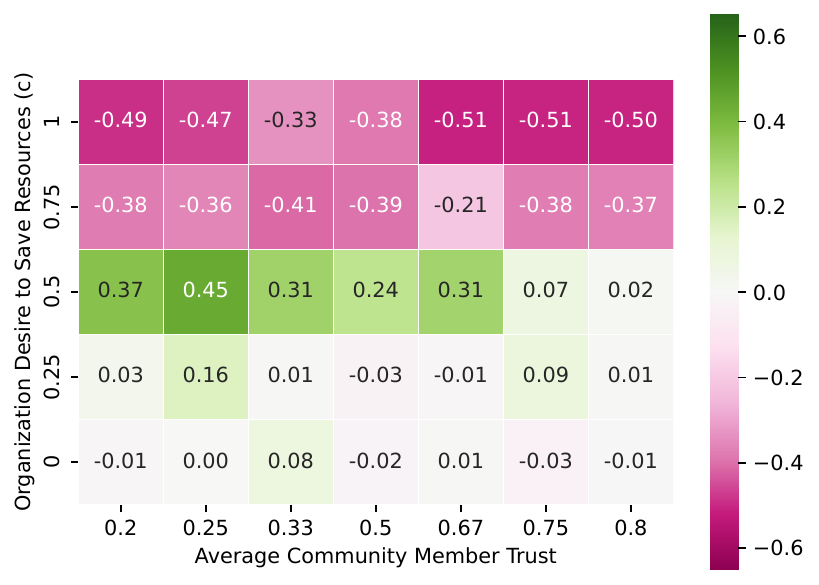} & \includegraphics[width = .32\linewidth]{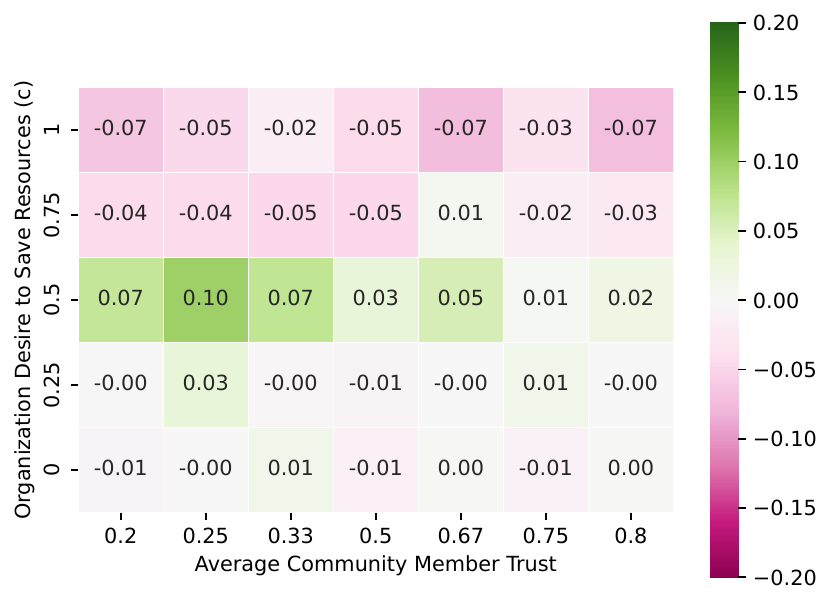}\\
    \small{(a) Quota $\mathcal{U}$ diff} & \small{(b) Quota fairness diff}  & \small{(c) Quota average trust diff}
    \end{tabular}
    \caption{Raw difference for each metric between our intervention and trust-aware RL algorithm. Green cells represent the interventions outperforming trust-aware, pink is the opposite, and white is neutral. \textbf{The organization's utility $\mathcal{U}$ suffers for high $c$ values, as does fairness and average community member trust. For mid to low $c$ values, fairness and average trust of agents increase.}}
    \label{fig:disc}
\end{figure}

We test this using the trust-aware model, and then compare the intervention results to those from the original trust-aware model in \Cref{fig:disc}. As is to be expected, the organization's utility $\mathcal{U}$ suffers for higher $c$ values, as the organization is explicitly trading off $|u|_{\gamma}$ to save resources. We see that to combat this, the organization attempts to allocate some resources to community members in the high $c$ cases, but does so unfairly, decreasing fairness and average trust. However, in the $c=0.5$ case, though $\mathcal{U}$ decreases, fairness and average trust of agents see a steady increase. The intervention is also successful in the case of $c=0.25$, benefiting both the organization and community. Here, we find that implementing a simple quota from an outside entity can have drastic effects on outcomes. 
We note here that though we assume the organization's reward function is static over time, one could explore the possibility of the organization dynamically changing its $c$ value in response to interventions. 
Future work could focus on studying such dynamic reward functions and designing optimized interventions, including collective action by community members.

\subsection{Operationalizing Trust}
In this work, we chose to model trust as a scalar, which is standard practice \cite{loi2023much}. And though the modeling choices we made were grounded in prior research \cite{cabiddu2022users, roy2015impact, yeager2017loss, qi2021perceived, lien2014trust, bhattacharya1998formal, mayer1995integrative, hancock2023and, granatyr2015trust}, our trust updates that could have been operationalized in different ways. For example, non-linear update rules or considering different factors in the trust updates are two feasible modeling changes. Beyond this, one might ask if we should be attempting to model trust at all. \citet{selbst2019fairness} argue that attempting to formalize complex human concepts often falls into the ``formalism trap,'' where definitions fail to account for the full meaning of these terms. However, in this work, we do not attempt to define institutional trust in its entirety---instead, we formalize institutional trust as serves us in this context. We recognize that we may not be capturing all aspects of institutional trust in this work, and the definition can be altered to reflect different priorities.

\subsection{Public Trust Values}
In \textbf{RQ1} we ask how an RL agent would learn a resource allocation policy if institutional trust for each community member were public---but how might we collect this data? We might think to survey community members, however we know that surveys are not accurate tools for capturing attitudes regarding trust \cite{miller2003surveys}. Since actually retrieving these values is difficult, it is essential that we ask and answer \textbf{RQ2}. 
The most likely scenario is that an institution will not be able to collect exact trust values, but rather must learn them from some signal. In our work, we use acceptance or rejection of services as the signal of trust. We recognize that in reality (not in our simulated environment), collecting enough data from this signal to make confident predictions would too be difficult.

\subsection{Impact of Network Structure}
Through our discussion of simulation results, which we presented in \Cref{sec:results}, we found that network structure had stronger affects on some metrics than others. While the utility to the organization ($\mathcal{U}$) was somewhat stable throughout the parameter space, we found larger fluctuations when it came to fairness and trust. These differences largely come from our definition of fairness, and other fairness metrics on networks could be explored as well \cite{liu2023group, de2024group}. Recall that the community members update their trust values partly due to the perceived fairness of their neighborhoods. Since the Gini coefficient measures income inequality, as the number of neighbors one community member has increases, it becomes more likely that the utilities observed by that community member are unequally distributed. Thus, as we see in \Cref{fig:rq1_ntwks}, isolated nodes maintain high trust values even when their accumulated service utility is low. Therefore, as larger network structures form, it is more difficult for the RL agent to learn policies that will not erode trust. 
%We further explore the impact of node degree in \Cref{sec:networks}.

\subsection{More Ethical Algorithmic Governance}
This work is predicated on the knowledge that algorithmic governance exists, and the belief that it will only become more prevalent as we come to rely more heavily on AI decision-makers. Our aim is not to make a case for or against algorithmic governance, but rather to point out one flaw in its current state, and propose a solution. We also demonstrate, through varying our organization's attribute $c$, or willingness of the organization to decrease service quality to save resources, that human decision-makers who design AI algorithms can have some control over fair outcomes.
This value $c$ may vary in reality for many reasons---a need to conserve resources for future use, or selfish desires of organization leaders to pocket any leftover budget. 
We acknowledge that exclusion of trust dynamics is not the only or even the most pressing issue with AI governance as it stands, but we hope that through this work we are able to make a small improvement.

In addition, we recognize that algorithms such as the one proposed here could be used by institutions with ill intent to harm communities. 
Specifically, we focus on attribute $c$, which as a reminder, is the organization's desire to save resources. We found that when the organization's desire to save resources is high, fairness and trust in the community suffer---thus, an organization that aims to harm communities could do so while still being successful by its own metric. We aim to avoid this outcome in our work. To this end, we design the quota system intervention (see \Cref{sec:quota}) to counteract the selfishness of the organization. In this work, we identify problems that could arise for communities, but we also design a solution. 

\section{Future Work and Limitations}
\textbf{Simulated data: }
Our work relies on simulated data and random assignment of some agent attributes, such as trust. 
This is a limitation of our work, though our simulated trust dynamics are designed carefully and grounded in prior research \cite{cabiddu2022users, roy2015impact, yeager2017loss, qi2021perceived, lien2014trust, bhattacharya1998formal, mayer1995integrative, hancock2023and, granatyr2015trust}. In an ideal setting, there would exist network datasets that describe the institutional trust of each node, with a governmental body that provides the community with some service. 
Unfortunately, as far as we are aware, one does not exist. We plan to collect such a dataset ourselves in the future, working with local organizations in order to interrogate the assumptions made in this paper. However, since the current work is an exploration of trust and how it affects RL algorithms, we find that aspect to be outside of the scope of this current work. 

\textbf{Trust updates: } 
We chose coefficients for trust updates so as to achieve a balance of fairness, current utility, and previous trust when updating the current trust value for each agent. Just as altering the value of $c$ in $\mathcal{U}$ impacted outcomes, changing the coefficients in the trust update may as well. For space, we chose not to include further experiments; instead, we prefer to conduct interviews with human subjects to better understand reasonable values for these coefficients, and interrogate whether these are the correct factors to include for a realistic trust update. However, future work could quite easily change the values of our modeling parameters.

\section{Conclusion}
In this work we developed a trust-aware reinforcement learning algorithm for resource allocation. We used DDPG to learn a resource allocation that maximizes the utility of the organization, and explored the impact on community fairness and institutional trust as that organization's utility function changed. We compared our trust-aware model to both trust-unaware and learned-trust, where the initial institutional trust values of community members are not known, but rather learned through other signals. This results from this learned-trust model are key when considering the difficulty an organization might face when attempting to actually elicit accurate trust values from community members. We found that most successful was in fact the learned-trust model---conservative Bayesian estimates were the best representation for institutional trust of community members. In the case where organizations prioritized saving resources over providing services to communities, the fairness and trust in those communities suffered. To combat these harmful outcomes, we explored a potential intervention, a quota system put in place by an external entity. We found that such a system improved fairness and trust in communities while decreasing the success of the organization. Further interventions could include collective action by citizens, and would make for interesting future work. 
As we examine this topic further, we plan in future work to conduct interviews to interrogate assumptions we made in this model, especially with respect to the data simulation and trust updates. 
In this work we explored one flaw in AI governance, though we recognize there are many. By including trust dynamics in RL simulations, we found that this addition could lead to more successful, fair, and trust-improving policies. 

%\begin{acks}

%\end{acks}

\bibliographystyle{ACM-Reference-Format}
\bibliography{ref}

%\appendix

%\input{sections/[99] appendix}

\end{document}